\documentclass{article}
\usepackage{ijcai24}

\usepackage{times}
\usepackage{soul}
\usepackage{url}
\usepackage[hidelinks]{hyperref}
\usepackage[utf8]{inputenc}
\usepackage[small]{caption}
\usepackage{graphicx}
\graphicspath{{figures/}}

\usepackage{amsmath}
\usepackage{amsthm}
\usepackage{amssymb}
\usepackage{booktabs}
\usepackage{algorithm}
\usepackage[noend]{algorithmic}
\usepackage[switch]{lineno}
\usepackage{makecell}
\usepackage{multirow}

\usepackage{svg}
\usepackage{courier}

\title{Integrating Intent Understanding and Optimal Behavior Planning for \\Behavior Tree Generation from Human Instructions}

\author{
Xinglin Chen$^{*}$ \and 
Yishuai Cai$^{*}$  \and 
Yunxin Mao\and 
Minglong Li$^\dag$ 
\and  \\
Wenjing Yang\and 
Weixia Xu\And 
Ji Wang
\affiliations
College of Computer Science and Technology, National University of Defense Technology\\
\emails
\{chenxinglin, caiyishuai, maoyunxin, liminglong10, wenjing.yang, xuweixia, wj\}@nudt.edu.cn
}

\begin{document}

\maketitle

\begin{abstract}

Robots executing tasks following human instructions in domestic or industrial environments essentially require both adaptability and reliability. Behavior Tree (BT) emerges as an appropriate control architecture for these scenarios due to its modularity and reactivity. Existing BT generation methods, however, either do not involve interpreting natural language or cannot theoretically guarantee the BTs' success. This paper proposes a two-stage framework for BT generation, which first employs large language models (LLMs) to interpret goals from high-level instructions, then constructs an efficient goal-specific BT through the Optimal Behavior Tree Expansion Algorithm (OBTEA). We represent goals as well-formed formulas in first-order logic, effectively bridging intent understanding and optimal behavior planning. Experiments in the service robot validate the proficiency of LLMs in producing grammatically correct and accurately interpreted goals, demonstrate OBTEA's superiority over the baseline BT Expansion algorithm in various metrics, and finally confirm the practical deployability of our framework. The project website is \href{https://dids-ei.github.io/Project/LLM-OBTEA/}{https://dids-ei.github.io/Project/LLM-OBTEA/}.

\end{abstract}

\section{Introduction}

Issuing instructions using natural language is a convenient and user-friendly way to control robots in everyday life scenarios, such as home, restaurants, and factories. However, enabling robots to autonomously follow these instructions remains a challenge in the field of robotics and AI. On the one hand, due to the openness of natural language, the robot must understand the user's intents, clarify the goals of tasks, and acquire sufficient high-level skills to accomplish these tasks. On the other hand, the robot must automatically generate reliable plans and execute them safely and robustly.

Behavior Trees (BTs) employ a logical tree structure to control various high-level robot behaviors, which offer many advantages, including modularity, interpretability, reusability, reactivity, and robustness. Due to these characteristics, BTs have become a popular robot control architecture in recent years. Moreover, approaches to automatically generate BTs have also been widely studied, including evolutionary computing \cite{colledanchise2019learning}, reinforcement learning \cite{banerjee2018autonomous}, and BT synthesis \cite{li2021reactive}. BT Expansion \cite{cai2021bt} is a sound and complete algorithm for constructing BTs through behavior planning, theoretically ensuring the success of the BT. However, existing methods typically generate BTs from fitness functions, reward functions, or formalized goal conditions, but few studies have focused on generating BTs directly from natural language instructions.

\begin{figure}[t]
	\centering
	\includegraphics[width=1 \columnwidth]{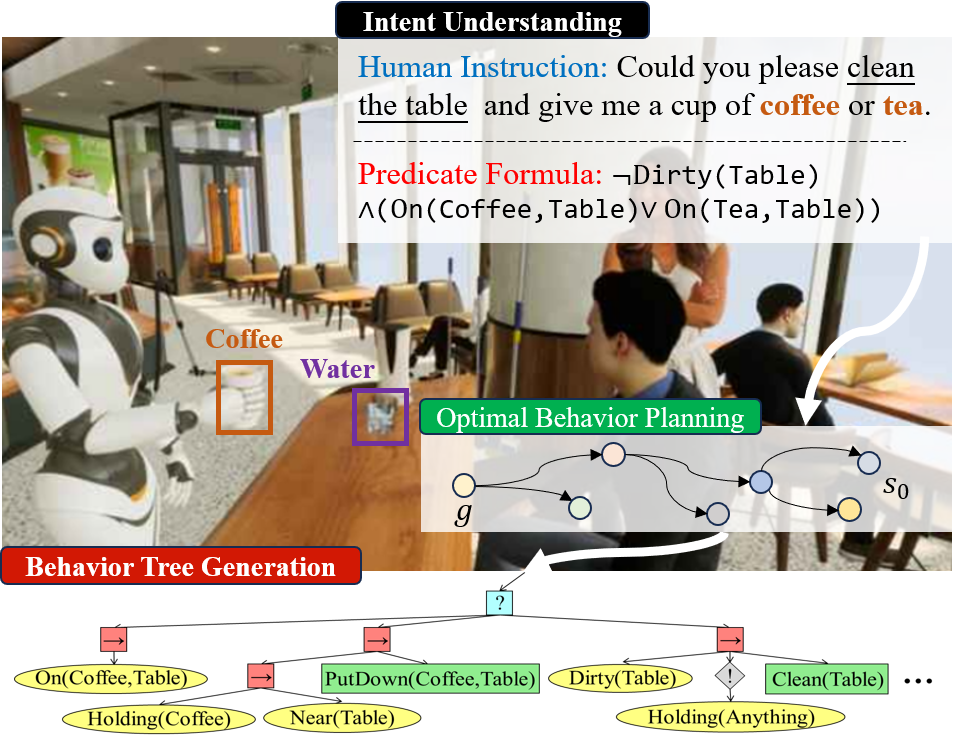}
	\caption{In our framework, human instructions are first understood and interpreted into a goal using LLMs. Then, OBTEA is employed to generate an optimal BT to achieve the goal. Goals are represented as well-formed formulas in first-order logic.}
	\label{fig:introduction}
\end{figure}

Large Language Models (LLMs), trained on extensive corpora to generate text sequences based on input prompts, exhibit remarkable multi-task generalization capabilities. This ability has recently been used to generate plausible task plans, which can be represented as action sequences \cite{song2023llmplanner}, codes \cite{singh2022progprompt}, and more. However, for the method of having LLMs directly generate BTs \cite{lykov2023llmbrain}, it not only necessitates gathering a large dataset of instruction-BT pairs for supervised fine-tuning, but also cannot guarantee theoretical correctness and optimality of the generated BTs.

This paper proposes a two-stage framework for efficient and reliable BT generation from human instructions. In the first stage, we utilize the LLM's capability in understanding intent to interpret a goal from the user's instructions. In the second stage, we design the Optimal Behavior Tree Expansion Algorithm (OBTEA) to construct a BT that theoretically can ensure the completion of the goal with minimal cost. The task goals, expressed as Well-Formed Formulas (WFFs) in first-order logic \cite{hodel2013introduction}, serve as a natural and ingenious bridge connecting these two stages. In a café scenario, for instance, there is a customer who says to the robot:  \textit{Could you please clean the table and give me a cup of coffee or tea?} The LLM generates a goal condition:
$G=$ {\ttfamily $\neg$Dirty$($Table$)\wedge$ $($On$($Coffee,Table$)\vee$On$($Tea,Table$))$}. Then, OBTEA traverses the action and condition space to find actions that can be used to achieve the goal, such as {\ttfamily Clean$($Table$)$} and {\ttfamily PutDown$($Coffee,Table$)$}. These actions are then structured to generate the final optimal BT, as shown in Figure \ref{fig:introduction}.

Our experiments demonstrate that the LLM (GPT 3.5), powered by few-shot demonstrations and reflective feedback, achieves high performance in both grammar and interpretation accuracy for intent understanding. The analysis of optimal behavior planning shows that the OBTEA produces BTs with lower costs and fewer condition node ticks compared to the baseline BT Expansion algorithm. Finally, the success of deployment in a café scenario highlights the promising future of our framework.

\begin{figure*}[t]
	\centering
	\includegraphics[width=1.96 \columnwidth]{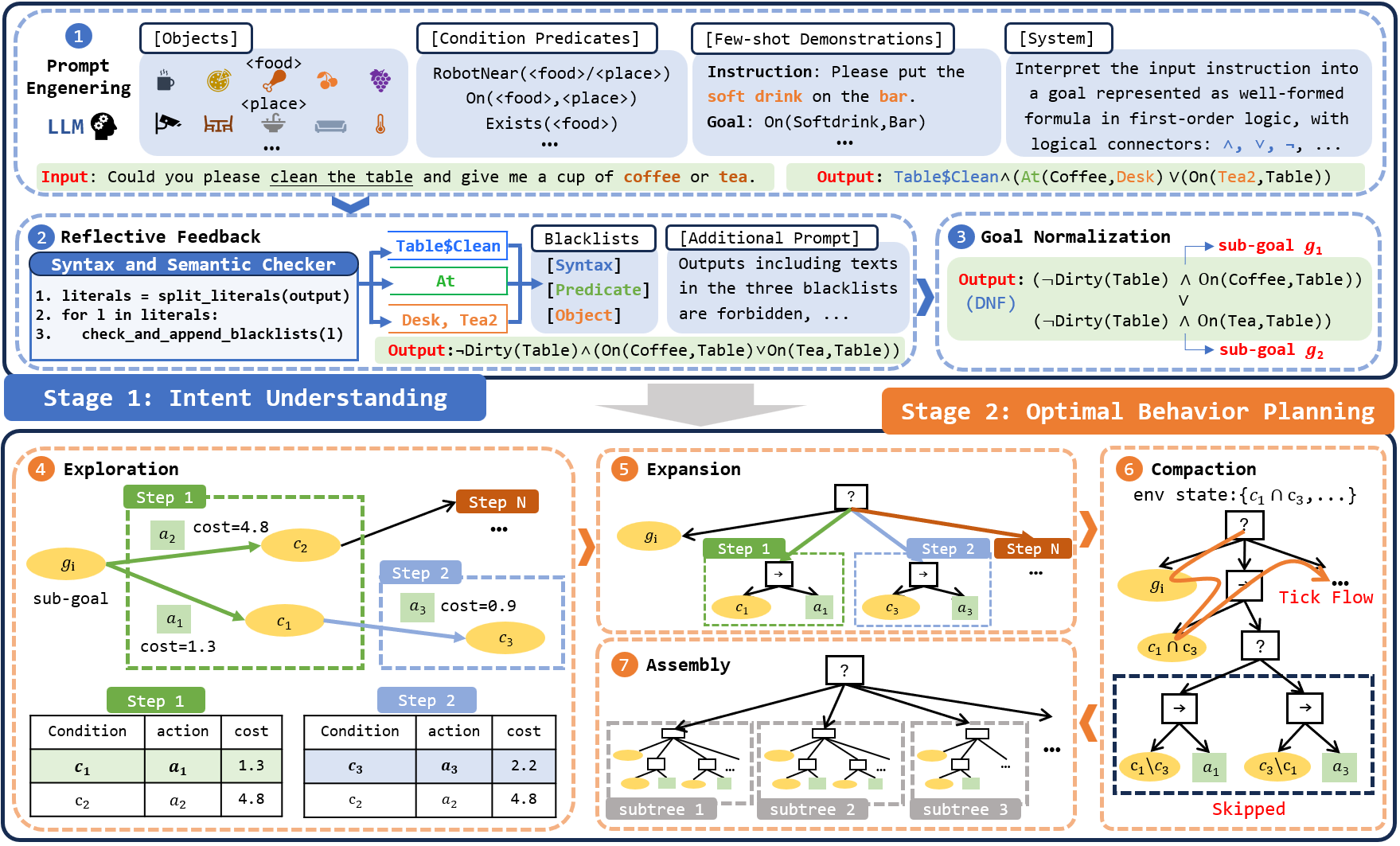}
	\caption{The two-stage framework. In Stage 1, the input instruction is transformed into a logically expressed goal by the LLM with prompt engineering and reflective feedback. The goal is then normalized as DNF and devided into sub-goals. In Stage 2, one subtree is generated for each sub-goal through exploration, expansion, and compaction. These subtrees are eventually assembled to create the final optimal BT.}
	\label{fig:framework}
\end{figure*}

\section{Preliminaries}

\paragraph{Behavior Tree.}
A BT is a directed rooted tree structure in which the leaf nodes control the robot's perception and actions, while the internal nodes handle the logical structuring of these leaves \cite{colledanchise2018behavior}. 
At each time step, the BT ticks from its root. The tick, according to the environment state, creates a control flow within the tree, ultimately determining the action that the robot will execute. In this paper, we focus on five typical BT nodes:

\begin{itemize}
 	\item \textbf{Condition}: 
 	A leaf node that evaluates whether the observed state satisfies the specified condition, returning either {\ttfamily success} or {\ttfamily failure} based on this assessment.
 	\item \textbf{Action}: A leaf node that controls the robot to perform an action, returning {\ttfamily success}, {\ttfamily failure}, {\ttfamily running} depending upon the outcome of execution.
 	\item \textbf{Sequence}: An internal node that only returns {\ttfamily success} if all its children succeed. Otherwise, it ticks its children from left to right, and the first child to return {\ttfamily failure} or {\ttfamily running} will determine its return status. 
 	It is often visualized as a $\rightarrow$ inside a box.
 	\item \textbf{Fallback}: An internal node with logic opposite to the sequence node. It returns {\ttfamily failure} only if all of its children fail. If not, the first occurrence of {\ttfamily success} or {\ttfamily running} during ticking becomes its return status.
 	It is often visualized as a ? in a box.
 	\item \textbf{Not-Decorator}: An internal node with a single child. It inverts the return statuses {\ttfamily success} and {\ttfamily failure} of its child. It is often visualized as a ! in a diamond.
 	
\end{itemize}

\paragraph{Behavior Planning.}
Based on \cite{cai2021bt} and \cite{colledanchise2018behavior}, we describe the behavior planning problem as a tuple: \(<\mathcal{S},\mathcal{A},P, s_0,g>\), where \( \mathcal{S} \) is the finite set of environment states, \( \mathcal{A} \) is the finite set of action, \( P \) defines the state transition rules, $s_0$  is the initial state, $g$ is the goal condition.  Both the state $s\in \mathcal{S}$ and the condition $c$ are represented by a set of literals, similar to STRIPS-style planning \cite{fikes1971strips}. If $c\subseteq s$, it is said that the condition $c$ holds in the state $s$. The state transition affected by the action $a\in \mathcal{A}$ can be defined as a triplet \( P(a)=<pre(a),add(a),del(a)> \), comprising the precondition, add effects, and delete effects of the action. An action $a$ can only be performed if $pre(a)$ holds in the current state $s$. After the execution of the action, the subsequent state $s'$ will be:
\begin{equation}
	s' = s \cup add(a) \setminus del(a)
\end{equation}

In well-designed BT systems, for each literal or its negation, there is a condition node to verify its truthfulness, and for each action, there is a corresponding action node for implementation. Given the initial state $s_0$ and the goal condition $g$, the objective of behavior planning is to construct a BT $\mathcal{T}$ capable of transitioning $s_0$ to $g$ in a finite time.

\paragraph{Problem Formulation.} In this paper, we extend the problem to optimal behavior planning, as the user's instructions often imply that the robot should complete the ordered task at minimal cost. Considering a well-defined cost function $ D $, where $\forall a\in \mathcal{A}, D(a)\geq 0$, for each BT \( \mathcal{T} \) produced by behavior planning, if the actions required to transition from \( s_0 \) to \( g \) are \( a_1, a_2, ..., a_n \), then the cost of the BT is calculated as \( D(\mathcal{T}) = \sum_{i=1}^{n} D(a_i) \). Our aim is to identify the optimal BT, \( \mathcal{T}_* \), which can achieve the goal condition $g$, as interpreted from human instructions, at the minimum cost.

\section{Methodology}

In this section, we first demonstrate how to represent goals as WFFs in first-order logic. Then, we dive into our two-stage framework for generating BT, where the input instruction is interpreted into the goal condition, and the OBTEA is applied to generate an optimal BT that guarantees to achieve the goal at the minimal cost. Our framework is illustrated in Figure \ref{fig:framework}. 

\subsection {Representing Goals as Well-Formed Formulas}

To describe the environment in first-order logic, we introduce a triplet $\langle \mathcal{O}, \mathcal{P}_c, \mathcal{P}_a \rangle$, where $\mathcal{O}$ is the set of available objects in the environment, $\mathcal{P}_c$ is the set of condition predicates, and $\mathcal{P}_a$ is the set of action predicates. For a predicate $p \in \mathcal{P}_c \cup \mathcal{P}_a$, its domain may be a subset of the Cartesian product of $\mathcal{O}$. In this formulation, a literal can be denoted as $p_c(o_1, ..., o_k)$ while an action can be denoted as $p_a(o_1, ..., o_l)$. Literals can be further formed into WFFs using three fundamental logical connectors: $\wedge$ (conjunction), $\vee$ (disjunction), and $\neg$ (negation). The three fundamental connectors effectively cover a wide range of user intentions, and more intricate connectors may be integrated in future work. This high-level formulation is advantageous for LLM to understand and interpret, as well as for the design and implementation of modular and reusable behavior nodes.

\begin{algorithm}[t]
	\caption{OBTEA}
	\label{alg:obtea}
	\textbf{Input}:  Goal $G$, init state $s_0$, action set $\mathcal{A}$, transition rules $P$ \\
	\textbf{Output}: The optimal BT \( \mathcal{T}_* \) 
	\begin{algorithmic}[1] 
		\STATE $\mathcal{G}_{sub} \gets ParseSubGoals(G)$ \label{line:initial} \hfill $\triangleright$ the set of sub goals
		\FOR{ $g_i \in \mathcal{G}_{sub}$ } \label{line:forSubGoal}
		\STATE $\mathcal{T} \gets Fallback(g_i)$ \label{line:initBT}
		\hfill $\triangleright$ init sub-BT
		\STATE $D(g_i)\gets 0$ \label{line:initCost}
		\hfill $\triangleright$ for $\forall c,c\neq g_i, D(c) \gets +\infty$
		\STATE $\mathcal{C}_U\gets \{g_i\}$ \label{line:cu}
		\hfill $\triangleright$ explored but unexpanded conditions
		\STATE $\mathcal{C}_E\gets \emptyset$ \label{line:ce}
		\hfill $\triangleright$ expanded conditions
		\WHILE{ $\mathcal{C}_U \neq \emptyset$}
		\STATE $c\gets \arg \min_{c\in\mathcal{C}_U} (D(c))$ \label{line:pickC}
		\hfill $\triangleright$ explore and expand $c$

		\FOR{\textbf{each} $a\in A$}   \label{line:fora}
		\IF{$(c\cap (pre(a)\cup add(a)\setminus del(a)) \neq \emptyset)$ and $(c\setminus del(a)=c)$} \label{line:forExploration}
		\STATE $c_{a}\gets pre(a) \cup c \setminus add(a)$ \label{line:exploration}
		\hfill $\triangleright$  \textbf{Exploration}
		\IF{ $c_a \notin \mathcal{C}_E$ and $D(c) + D(a) < D(c_{a})$ } \label{line:ifExplored}
		\STATE $D(c_{a})\gets D(c) + D(a) $   \label{line:updateD}
		\hfill $\triangleright$ update $D(c_a)$
		\STATE $\mathcal{C}_U\gets \mathcal{C}_U\cup \{c_{a}\}$
		\hfill $\triangleright$ add $c_a$ to $\mathcal{C}_U$
		\STATE $\mathcal{M}(c_{a})\gets Sequence(c_a,a)$   
		\label{line:storePair}
		\ENDIF
		\ENDIF
		\ENDFOR
		
		\STATE $\mathcal{C}_U\gets \mathcal{C}_U\setminus \{c\}$ \label{line:removeC}
		\hfill $\triangleright$ remove $c$ from  $\mathcal{C}_U$

		\IF{$c\neq g_i$}
		\STATE $\mathcal{T} \gets Fallback(\mathcal{T},\mathcal{M}(c))$ \label{line:expand} 
		\hfill $\triangleright$ \textbf{Expansion}
		\IF{ $c \subseteq s_0$} \label{line:expandS0}
		\STATE $D(\mathcal{T})\gets D(c)$ \label{line:totalCost}
		\hfill $\triangleright$ total cost of $\mathcal{T}$
		\STATE \textbf{break}  \label{line:break}
		\hfill $\triangleright$  $\mathcal{T}$ is the optimal BT for sub-goal $g_i$
		
		\ENDIF
		\ENDIF
		\STATE $\mathcal{C}_E\gets \mathcal{C}_E\cup \{c\}$ \label{line:addC}
		\hfill $\triangleright$ add $c$ to $\mathcal{C}_E$

		\ENDWHILE
		\STATE $\mathcal{T}_i \gets Compact(\mathcal{T})$  \label{line:compaction} 
		\hfill $\triangleright$  \textbf{Compaction}
		\ENDFOR
		\STATE $\mathcal{T}'_1,\mathcal{T}'_2,...,\mathcal{T}'_{|\mathcal{G}_{sub}|} = \underset{key=D(\mathcal{T})}{SortAscending} (\mathcal{T}_1,\mathcal{T}_2,...,\mathcal{T}_{|\mathcal{G}_{sub}|})$ \label{line:sort} 
		\STATE $\mathcal{T}_* \gets Fallback(\mathcal{T}'_1,\mathcal{T}'_2,...,\mathcal{T}'_{|\mathcal{G}_{sub}|})$ \label{line:assembly} 
		\hfill $\triangleright$ \textbf{Assembly}
		\RETURN $\mathcal{T}_*$
		
	\end{algorithmic}
\end{algorithm}

\subsection{Intent Understanding}

Given the relatively interpretable nature of first-order logic grammar in everyday service scenarios, we believe that pre-trained LLMs have the potential to interpret the instruction without supervised fine-tuning on specialized datasets. We achieve this process through the following three methods.

\paragraph{Prompt Engineering.} 
We deliberately design LLM prompts to provide context for goal generation, similar to \cite{singh2022progprompt,ahn2022can}. Our prompts consist of four components: (1) Objects. The set of available objects. To better define the domains of predicates, we categorize objects into several classes, such as  {\ttfamily $<$food$>$} and  {\ttfamily <place>}. (2) Condition Predicates. The set of condition predicates. The domain of a predicate is indicated by category, such as {\ttfamily RobotNear$($<place>$)$} and {\ttfamily On$($<food>,<place>$)$}. (3) 
Few-shot Demonstrations. A few instruction-goal pairs as examples to teach the LLM on goal interpretation. The number of demonstrations does not need to be large, but it is beneficial to showcase as much diversity in both grammar and user's intention as possible. (4) System. An explanatory text provides context and the prompt for instructing the LLM to produce a goal in the correct format. We found that few-shot demonstrations have a significant impact on the LLM's ability to interpret goals, as language models have been shown to be few-shot learners \cite{brown2020language}. We conducted ablation experiments in Section \ref{sec:few-shot} to analyze its impact.

\paragraph{Reflective Feedback.}

We introduce automatic reflective feedback to further enhance grammar accuracy, drawing inspiration from \cite{liu2023fimo}. When the LLM produces an output, the syntax checker evaluates whether the output forms a WFF, while the semantic checker further verifies the validity of predicates and objects. If errors are detected, we re-input the instruction to the LLM with additional prompt to correct these errors. While we initially provide whitelists of objects and condition predicates in the original prompt, we augment the prompt with blacklists and additional context to instruct the LLM not to include the blacklisted texts in the output again. If errors persist, we continue to apply reflective feedback, gradually expanding the blacklists, until all grammar errors are corrected, or the maximum retry limit is reached.

\paragraph{Goal Normalization.} 

A grammatically correct goal will ultimately undergo normalization into Disjunctive Normal Form (DNF). This normalization offers two significant advantages. Firstly, it effectively eliminates ambiguity from WFFs, which is crucial when assessing the equivalence of two goals, particularly in evaluating the semantic correctness of the model's outputs on the evaluation dataset. Secondly, each conjunction within DNF can be regarded as a sub-goal. When the robot accomplishes any one of these sub-goals, it is considered a success, as these sub-goals are represented as disjunctive relations. This enables us to design a modular algorithm for optimal behavior planning.

\subsection{Optimal Behavior Planning}

We introduce the OBTEA for optimal behavior planning based on the goal \( G \) represented in DNF (Fig. \ref{fig:framework} and Alg. \ref{alg:obtea}).
At the beginning of OBTEA, we parse the goal \( G \) to identify sub-goals separated by \( \vee \). Each sub-goal is then further divided by \( \wedge \) to form a set of literals, each consisting of a goal condition \( g_i \). This process results in a set of goal conditions that constitute the sub-goal set \( \mathcal{G}_{sub} \) (line \ref{line:initial}). For example, if the goal in DNF is \( G = (l_1 \wedge l_2) \vee (l_1 \wedge l_3) \), this translates to two sub-goals: \( l_1 \wedge l_2 \) and \( l_1 \wedge l_3 \). Consequently, the sub-goal condition set becomes \( \mathcal{G}_{sub} = \{ g_1, g_2 \} \), where \( g_1 = \{ l_1, l_2 \} \) and \( g_2 = \{ l_1, l_3 \} \). Note that we facilitate more flexible human instructions by allowing negative literals \( \neg l \), which are often banned in action planners for performance reasons. Users must weigh the trade-off between interpretability and computational complexity in practical applications.

For each sub-goal \( g_i \in \mathcal{G}_{sub} \), we construct an optimal subtree (line \ref{line:forSubGoal}). Initially, each subtree is rooted in a fallback node, with its child node being the sub-goal condition \( g_i \) (line \ref{line:initBT}). To track the shortest distance, we maintain a cost function \( D \), which is initialized with \( D(c_g) = 0 \) and \( D(c) = +\infty, \forall c\neq g_i  \) (line \ref{line:initCost}). Furthermore, we maintain two sets: \( \mathcal{C}_U \) and \( \mathcal{C}_E \). \( \mathcal{C}_U \) encompasses explored but unexpanded conditions, initialized with \( \{g_i\} \) (line \ref{line:cu}). \( \mathcal{C}_E \) includes conditions that have been expanded, initially empty (lines \ref{line:ce}).

\paragraph{Exploration.}

We continuously pop the condition $c\in \mathcal{C}_U$ with the minimum cost (line \ref{line:pickC}). This process aims to explore the condition space following the shortest path from $g_i$. For each popped condition $c$, we explore its neighboring conditions by iterating through all available actions (line \ref{line:fora}) and identifying those that can reach the condition $c$ after execution. The neighboring conditions are generated similarly to the process in the BT Expansion \cite{cai2021bt} (lines \ref{line:forExploration}-\ref{line:exploration}), and can be denoted as:
$c_{a}\gets pre(a) \cup c \setminus add(a)$. Then we disregard $c_a$ in the following two cases (line \ref{line:ifExplored}): (1) $c_a \in \mathcal{C}_E$, indicating that it has already been expanded and its shortest path is found. (2) $D(c) + D(a) \geq D(c_{a})$, showing that $c_a$ has been explored and the path cost found is equal or lower. In both cases, further exploration or updates of $c_a$ are unnecessary. If neither case applies, we explore $c_a$ by updating $D(c_a)$, adding $c_a$ to $\mathcal{C}_U,$ and recording the subtree $Sequence(c_a,a)$ leading to $c_a$ (line \ref{line:updateD}-\ref{line:storePair}). After exploring all the neighbors of $c$, it is removed from $\mathcal{C}_U$ (line \ref{line:removeC}).

\paragraph{Expansion.} 

After exploring $c$, we integrate it into the current BT $\mathcal{T}$. This involves adding the subtree $\mathcal{M}(c)$ as a child node of the root fallback node (line \ref{line:expand}). 
The algorithm continues to explore the condition space until it expands a condition \( c \) that holds in the initial state \( s_0 \) (line \ref{line:expandS0}). At this point, we record the total cost of the subtree for \( g_i \) as \( D(\mathcal{T}) = D(c) \) (line \ref{line:totalCost}), and the exploration-expansion loop is immediately terminated (line \ref{line:break}). If \( s_0 \) is not reached, we add \( c \) to the expanded set \( \mathcal{C}_E \) (line \ref{line:addC}). It is important to note that this method of expansion might include conditions that are not part of the shortest path. However, this does not affect the optimality since these extra conditions will not be visited along the shortest path. This approach enhances the BT's reactivity and robustness when the state deviates from the original path, ensuring that the path starting from each expanded condition is optimal.

\begin{table*}[t] 

	\resizebox{\linewidth}{!}{
		\huge
		\renewcommand\arraystretch{1.1}
			\begin{tabular}{|c|c|c|c|c|c|c|c|c|}  
                    \hline
				Difficulty & Data & Examples & Goals & Demonstrations & GA-NF(\%) & GA-1F(\%) & GA-5F(\%) & IA(\%) \\ 
				\hline
    
				\multirow{3}{*}{Easy} & \multirow{3}{*}{30} & \multirow{3}{*}{\makecell[l]{Please come to the bar. }} & \multirow{3}{*}{{\ttfamily RobotNear(Bar)}} 
			& Zero-Shot & 60.0 & 74.7 & 91.3 & 56.7 \\
			& & & & Few-shot-1 & 87.3 & 95.3 & 100.0 & 85.3 \\
			& & & & Few-shot-5 & 92.0 & 96.7 & \textbf{100.0} & \textbf{90.7} \\ 
                    \hline
				
				\multirow{3}{*}{Medium} & \multirow{3}{*}{30} & \multirow{3}{*}{\makecell[l]{Ensure there's enough water\\ and keep the hall light on. }} & \multirow{3}{*}{\makecell[l]{{\ttfamily Exist(Water)} $ \wedge$ \\ $ $ {\ttfamily Active(HallLight)}}}
			& Zero-Shot & 50.0 & 66.0 & 88.0 & 57.3  \\
			& & & & Few-shot-1 & 77.3 & 91.3 & 98.7 & 62.0  \\
			& & & & Few-shot-5 & 92.7 & 94.7 & \textbf{99.3} & \textbf{82.0} \\  
                    \hline
				\multirow{3}{*}{Hard} & \multirow{3}{*}{40} & \multirow{3}{*}{\makecell[l]{I'm at table three, please bring \\yogurt and keep the AC warm. \\ If there is no yogurt, coffee is fine.}} & \multirow{3}{*}{\makecell[l]{ {\ttfamily (On(Yogurt,Table3)} $ \wedge$ \\ {\ttfamily(On(Coffee,Table3))} $\vee$ \\ $\neg$ {\ttfamily Low(ACTemperature)}}} 
			& Zero-Shot & 48.0 & 62.5 & 84.0 & 42.0  \\
			& & & & Few-shot-1 & 76.5 & 87.5 & 93.0 & 51.0  \\
			& & & & Few-shot-5 & 80.5 & 88.0 & \textbf{96.0} & \textbf{65.0}  \\
                    \hline
			\end{tabular}
	}
    \caption{Performance analysis of GPT 3.5 on intent understanting in the café sencario.}
	\label{table:LLMresult}
\end{table*}

\paragraph{Compaction.}
For an expanded optimal subtree, we propose a simple but efficient method to compact its structure and improve runtime efficiency. During compaction, we recursively extract common literals from two adjacent conditions to create smaller conditions for checking. Consider a BT produced through the exploration and expansion process. If two adjacent subtrees have conditions $c_1$ and $c_3$, respectively (as illustrated in Fig. \ref{fig:framework}), both conditions need to be checked during runtime. After compaction, if $c_1\cap c_3 \neq \emptyset$, a new condition $c'=c_1\cap c_3$ will be created and checked first during ticking. If it does not hold, the conditions $c_1 \setminus c_3$ and $c_3 \setminus c_1$ will be skipped, thereby saving time. The compacted subtrees can recursively join the compaction with its neighbors due to the modularity of BT. We evaluated the impact of the maximum recursion depth during compaction on condition check count in Section \ref{sec:exp-obtea}.

\paragraph{Assembly.}

Once all subtrees have been generated, we can assemble the final optimal BT. We start by sorting all the subtrees in ascending order according to their total costs (line \ref{line:sort}). This ensures that the BT will prioritize the feasible paths with the lowest cost. Then, we construct the final BT $\mathcal{T}_*$ by adding the sorted subtrees under a fallback node (line \ref{line:assembly}). After this, the algorithm ends and returns $\mathcal{T}_*$ as the optimal BT to achieve the final goal $G$ at minimal cost.

\subsubsection{Evaluation}

The BT generated by OBTEA is theoretically guaranteed to be finite-time successful and optimal, assuming that all sub-goals can be reached from the initial state $s_0$ within the BT system. Similar to BT Expansion \cite{cai2021bt}, the time complexity of OBTEA is also $O(|\mathcal{A}||\mathcal{S}|\log(|\mathcal{S}|))$ in the worst case, which is polynomial in terms of the system size. Proofs and a detailed discussion are in the appendix.

\section{Experiments}
\subsection{Settings}

We conducted experiments and deployed our framework in a hospitality industry scenario, where the robot plays the role of a waiter in a café. It is responsible for serving customers and performing various tasks following human instructions. In this scenario, the design of object and action sets is similar to the Virtual Home \cite{puig2018virtualhome} and Alfred \cite{shridhar2020alfred} benchmarks, but it is tailored to better suit the specific requirements of a café. The scenario includes a total of 80 available objects, 8 condition predicates: $\mathcal{P}_c=\{${\ttfamily RobotNear,Holding,On,Closed,Exists,Dirty
,Activate,Low}$\}$ and 6 action predicates $\mathcal{P}_a=\{${\ttfamily Make,
MoveTo,PickUp,PutDown,Clean,Turn}$\}$. 
Different predicates have different domains and implementations on behavior nodes, resulting in a total of 1318 extendable literals and 1269 extendable actions.

\subsection{Analysis for Intent Understanding}

To evaluate the intent understanding capabilities of LLM for goal interpretation, we created an evaluation dataset comprising 100 instructions and their corresponding expected goals, categorized into three difficulty levels. As exampled in Table \ref{table:LLMresult}, the Easy set consists of goals formed by a single positive literal. In the Medium set, goals consist of two literals connected by $\vee$ or $\wedge$. The Hard set involves three literals that incorporate $\vee$, $\wedge$, and $\neg$. The performance of the LLM in the interpretation of goals is evaluated primarily based on two metrics: grammatical accuracy (GA) and interpretation accuracy (IA). GA is the fraction of output goals that are free from grammatical errors. IA is the fraction of goals that are equivalent to the ground truth. 

Our methods are model-agnostic, allowing for application to any LLM. We conducted the experiments with GPT-3.5, specifically using the gpt-3.5-turbo model (2024.06). Table \ref{table:LLMresult} presents the performance of the model in three levels of difficulty, with each result being the average of 5 runs. We studied the impact of two factors on our method: the number of demonstrations and the maximum retry limits. Our standard setting involves five demonstrations with a limit of five retries for reflective feedback. At the easy level, we find the LLM performance is surprisingly high, with GA and IA reaching 100$\%$, 99.3$\%$ and 96$\%$, respectively. At medium and hard levels, the decrease in IA is faster than in GA. This suggests that the grammar of first-order logic is easier for LLMs to learn, but the translation from natural language is more challenging. Overall, both GA and IA are relatively high even at the hard level, demonstrating the LLM's potential for interpreting WFFs even without any supervised fine-tuning.

\begin{table}[t]
	\centering
	\small
	\renewcommand{\arraystretch}{1}
	\begin{tabular}{c|cc|cc}
		\toprule
		\multirow{2}{*}{Case} & \multicolumn{2}{c|}{Total Costs} & \multicolumn{2}{c}{ Condition Node Ticks} \\  
		
		& Baseline & OBTEA & Baseline & OBTEA  \\ 
		\midrule
		Easy & 123.7 & \textbf{67.4} & 65.4 & \textbf{14.2}  \\
		Medium & 171.1 & \textbf{92.9} & 271.1 & \textbf{105.0} \\
		Hard & 204.8 & \textbf{111.7} & 1437.4 & \textbf{809.7}  \\
		\bottomrule
	\end{tabular}
     \caption{Performance comparison on optimal behavior planning in the café scenario.}
     \label{tab:Experiment-obp-cafe} 
     
\end{table}

\begin{table*}[t]
	\centering

	\label{tab:Experiment_total}
	\small
	\renewcommand{\arraystretch}{1}
	\begin{tabular}{c|cccc|ccc|cc|ccc}
		\toprule
		
		\multirow{2}{*}{Case} & \multicolumn{4}{c|}{Scenario Configuration} & \multicolumn{3}{c|}{Problem} &  \multicolumn{2}{c|}{Total Costs} &  \multicolumn{3}{c}{Condition Node Ticks} \\  
		
		& \makecell{$|\mathcal{O}|$} & \makecell{$|\mathcal{P}_c|$} & \makecell{$|\mathcal{P}_a|$} & \makecell{MAC} & Literals & States & Actions & Baseline & OBTEA  & Baseline & OBTEA & OBTEA-NC \\ 
\midrule
		
		0 & 100 & 10 & 10 & 0 & 995.4 & 475.5 & 509.5 & 71.3 & \textbf{58.5} & 31.7 & \textbf{23.8} & 33.1     \\
1 & 100 & 10 & 50 & 0 & 1000.0 & 4046.5 & 2495.9 & 52.0 & \textbf{26.8} & 25.7 & \textbf{22.7} & 33.8     \\
2 & 500 & 50 & 50 & 0 & 25000.0 & 22468.6 & 12508.2 & 164.2 & \textbf{137.0} & 950.4 & \textbf{632.3} & 1779.6     \\
3 & 100 & 10 & 10 & 5 & 995.1 & 466.1 & 527.3 & 71.7 & \textbf{26.1} & 54.0 & \textbf{28.5} & 54.7     \\
4 & 100 & 30 & 10 & 5 & 2984.4 & 501.0 & 530.4 & 170.9 & \textbf{81.1} & 597.1 & \textbf{358.2} & 1005.1     \\
5 & 100 & 50 & 10 & 5 & 4974.6 & 497.7 & 524.4 & 241.4 & \textbf{124.8} & 1581.2 & \textbf{1210.5} & 3578.8     \\
6 & 100 & 50 & 30 & 5 & 5000.0 & 2496.0 & 1527.4 & 189.4 & \textbf{97.7} & 1391.9 & \textbf{1147.9} & 3111.3     \\
7 & 100 & 50 & 50 & 5 & 5000.0 & 4501.6 & 2531.5 & 169.4 & \textbf{85.1} & 1584.4 & \textbf{971.8} & 2848.0     \\
8 & 300 & 50 & 50 & 5 & 15000.0 & 13550.7 & 7560.9 & 164.2 & \textbf{82.5} & 1302.4 & \textbf{892.8}  & 2577.2    \\
9 & 500 & 50 & 50 & 5 & 25000.0 & 22427.9 & 12514.2 & 165.9 & \textbf{82.7} & 1436.6 & \textbf{909.8} & 2581.2     \\
		\bottomrule
	\end{tabular}
    \caption{Performance comparison on behavior planning in the computational test sets. Each values are averaged over 1000 runs.}
  \label{tab:bt compare result}
\end{table*}

\paragraph{Ablating Few-shot Demonstration.}
\label{sec:few-shot}\label{sec:feedback}

To conduct the ablation analysis on the number of demonstrations, we reduced the number of demonstrations while keeping the objects, predicates, and system explanations constant. It is observed that when no demonstrations are provided (zero-shot), both GA and IA significantly decrease across three levels of difficulty. Furthermore, IA is particularly lower than GA in the zero-shot setting, which again confirms the challenge for LLM to learn intent understanding rather than learning logical grammar only in explanatory prompts. Without any demonstrations, the pre-trained LLM struggles to provide WFF-represented goals that align with the user's intent. However, there is a notable improvement in performance after adding just one demonstration (one-shot). This phenomenon indicates that one example can be more effective than a lengthy explanation. The results underscore the importance of including examples in prompts and also highlight the LLM's few-shot learning ability in interpreting goals.

\paragraph{Ablating Flextive Feedback.}

We then performed the ablation analysis on the maximum retry times with reflective feedback. The results show that without feedback (GA-NF), the performance is markedly decreased, especially in the absence of demonstrations. Similarly to the number of demonstrations, GA improved when the LLM is allowed to retry only once (GA-1F), and an increase in the number of feedback times continuously improved grammar correctness. These results confirm the effectiveness of reflective feedback.

\subsection{Analysis for Optimal Behavior Planning}
\label{sec:exp-obtea}

Once the LLM produces the correct goal, OBTEA is guaranteed to generate an optimal BT to achieve that goal. We evaluated the BT generation performance for 100 ground-truth goals in the café scenario, where the action cost is set based on human experience. We focus on two key metrics: total costs of the BT and the condition node ticks during the BT's execution. Table \ref{tab:Experiment-obp-cafe} shows the performance comparison between OBTEA and the baseline BT Expansion algorithm \cite{cai2021bt}. The results demonstrate that our OBTEA generates BTs with lower costs and fewer condition node ticks than the baseline, and thereby has higher execution efficiency.

\paragraph{Random Test Sets.} In addition to the dataset in the café scenario, we also designed random test sets to verify the generality of OBTEA.
Inspired by \cite{cai2021bt}, we randomly generate various optimal behavior planning tasks by creating action sequence paths from the initial state to the goal and then randomly adding other actions. In contrast to previous work, our datasets are represented using first-order logic. The problem scale is mainly determined by the number of objects ($\mathcal{O}$), condition predicates ($\mathcal{P}_c$), and action predicates ($\mathcal{P}_a$). To increase the number of valid paths to the goal, we can randomly copy the actions in the original path several times, but assign different costs, denoted as maximum action copies (MAC). Further details can be found in the appendix.

Table \ref{tab:bt compare result} shows the results averaged over 1000 runs. The results again emphasize OBTEA's proficiency in achieving lower action costs and condition node ticks across all cases with varying complexity. In particular, in the most complex case 9, OBTEA significantly reduces costs by more than $50\%$, and condition node ticks decrease by approximately $30\%$. When there are more available paths to reach the goal, the costs of BTs found by OBTEA are smaller than the baseline. Note that while OBTEA has achieved improvements, its additional time overhead is also acceptable. The test results and discussions regarding planning time are in the appendix.

\paragraph{Ablating Compaction.}

\begin{figure}
	\centering
	\includegraphics[width=0.88 \columnwidth]{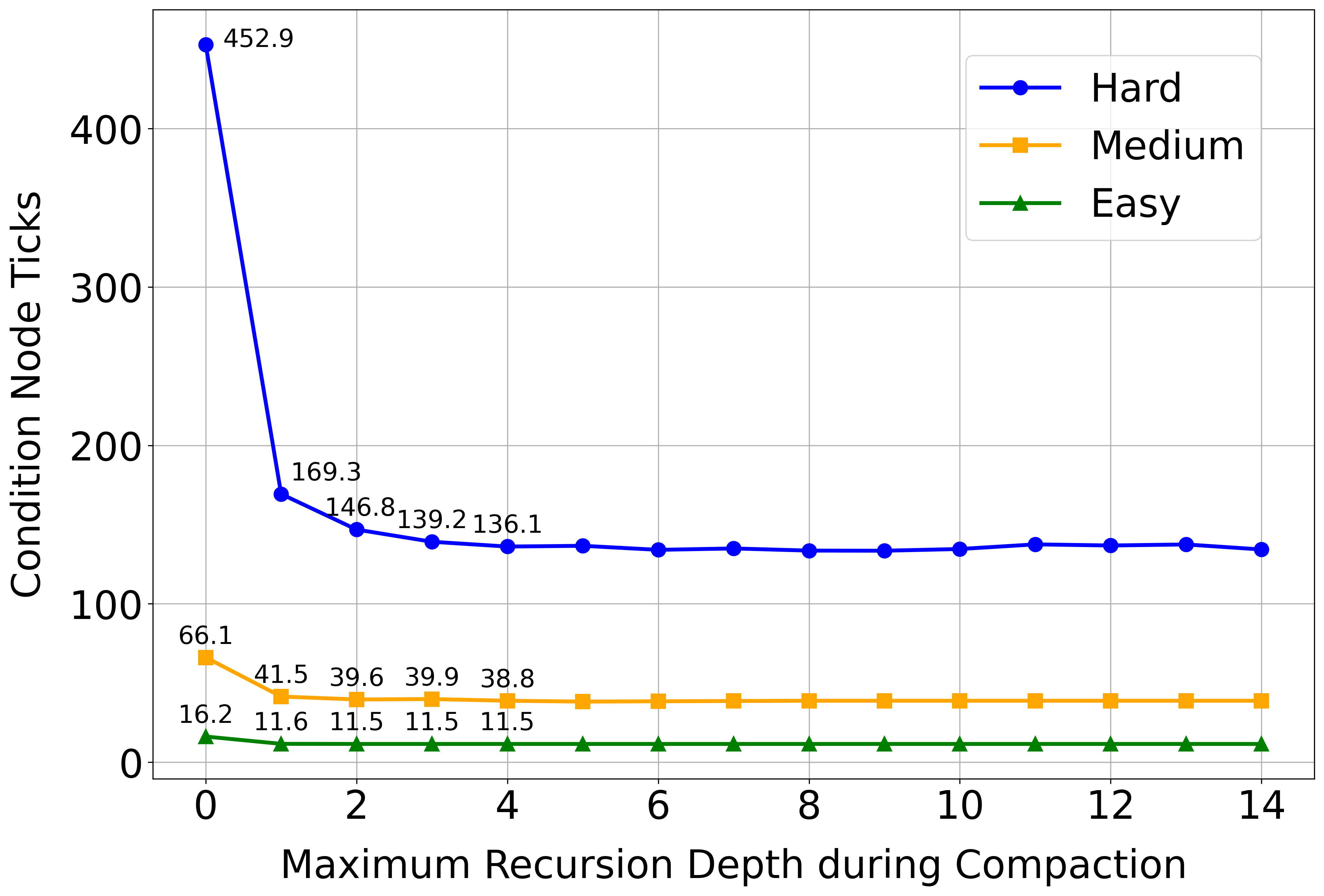}
	\caption{The impact of maximum recursion depth on condition node ticks during compaction in the café scenario.}
	\label{fig:merge_result}
\end{figure}

We then performed an ablation analysis on the impact of the maximum recursion depth during compaction. Table \ref{tab:bt compare result} shows that when we remove compaction (OBTEA-NC), the condition node ticks increase significantly in all cases. Experiments in café scenarios (Figure \ref{fig:merge_result}) show that condition node ticks decrease rapidly as the recursion depth increases when the recursion depth is small and then become rather constant when we continuously increase the depth. This means that the recursion depth does not need to be set too large to achieve optimal performance in the café scenario.

\begin{figure*}
	\centering
	\includegraphics[width=2.05 \columnwidth]{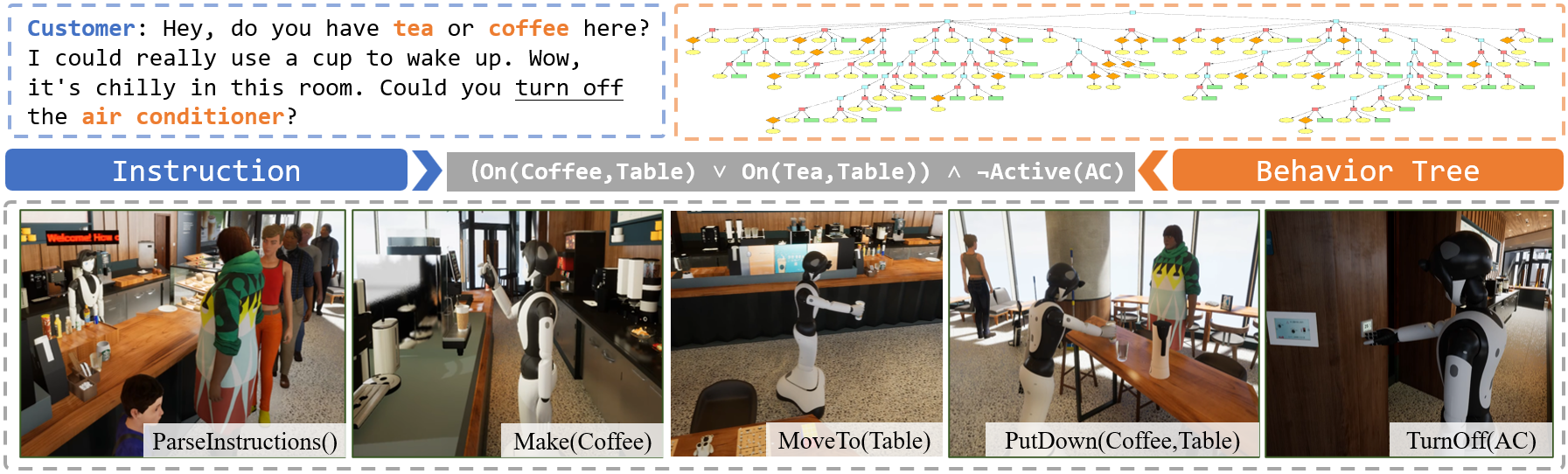}

	\caption{An example of the robot executing tasks follows the customer's instruction. After receiving the instruction at the bar, the robot generates an optimal BT through intent understanding and optimal behavior planning. The robot then autonomously completes the goal under the BT's control, which involves making the coffee, delivering it to the customer, and turning off the air conditioner. }
	\label{fig:development}

	\label{fig:development}
\end{figure*}

\subsection{Deployment}

We implemented our framework in a digitally twinned café environment constructed using the simulator used in MO-VLN \cite{liang2023movln}, as shown in Figure \ref{fig:development}. In the simulator, a humanoid robot acts as a waiter and has 21 active joints for precise movement control. We implement 8 condition nodes and 6 action nodes, corresponding to conditions and action predicates. To gather object information, the robot initially performs visual segmentation and detection to determine object poses from multiple perspectives. For efficient condition checks, the robot actively explores and continuously updates a semantic topological map for task execution queries.
By integrating these technologies, we ultimately achieve an adaptable, comprehensive, and reliable café waiter robot system capable of visual-language navigation, manipulation, and handling complex combination tasks. Our work was presented and won the First Prize in a competition at ChinaSoft 2023.
The successful deployment demonstrates the utility and practicality of our framework, representing a promising behavior tree generation framework in embodied intelligence scenarios.

\section{Related Work}

\paragraph{BT Generation.} The study of automatic BT generation is a significant area of research in robotics and AI. Various methods have been explored for this purpose. Heuristic search methods, such as grammatical evolution \cite{neupane2019learning}, genetic programming \cite{colledanchise2019learning,lim2010evolving}, and Monte Carlo DAG Search \cite{scheide2021behavior}, have been widely investigated. 
In addition to heuristic approaches, machine learning techniques have been applied to generate BTs. In particular, methods such as reinforcement learning \cite{banerjee2018autonomous,pereira2015framework} and imitation learning \cite{french2019learning} have gained attention. 
There are also research efforts that employ formal synthesis methods to generate BTs, such as Linear Temporal Logic (LTL) \cite{li2021reactive} and its variant approaches \cite{tadewos2022specificationguided,neupane2023designing}.
However, it is worth noting that this approach can be resource intensive in terms of modeling the environment and the computational resources required to solve the problem. 

\paragraph{LLMs for Task Planning.} LLMs have gradually gained widespread application in the field of robot task planning recently.
Some research focuses on directly controlling the low-level actions of robots using multi-modal language models \cite{driess2023palme}, while others use large language models to generate high-level behavior planning for robots. 
Generating BTs based on human instructions is an area that has seen limited exploration. One straightforward approach is to train a large language model to generate behavior tree structures represented in XML format \cite{lykov2023llmbrain,lykov2023llmmars}.
This approach is similar to the methods where LLMs directly generate action sequences \cite{song2023llmplanner,liu2023llm,ahn2022can,chen2023robogpt}, or codes \cite{liang2022code,singh2022progprompt,zeng2022socratic}.
However, these approaches may not fully leverage the modularity and computability of BTs for planning, potentially resulting in a loss of the reliability and interpretability advantages that BTs offer.

Our approach combines the ability of LLMs to understand intent with first-order logical reasoning for correctness assurance, ultimately achieving BT generation based on human instructions.
This approach aligns with the concepts of neurosymbolic approaches \cite{chaudhuri2021neurosymbolic} and symbolic planning \cite{liu2023fimo}. 
In the future, it is possible to enhance the generalization of robot planning in new scenarios by combining techniques like rule discovery \cite{zhu2023large}.
Hence, our framework is considered a promising approach for creating interpretable and reliable embodied intelligent agents.

\section{Conclusion}
This paper proposes a two-stage framework for BT generation, leveraging the strengths of LLMs for intent understanding and introducing the OBTEA for optimal behavior planning. The goal is represented as a well-formed formula in first-order logic, serving as a natural and innovative link between these two stages. Our framework integrates intent understanding and optimal behavior planning, which shows promise in creating adaptable and reliable generative embodied agents. In addition, we deployed our framework in a café scenario, showcasing its significant potential.

In the future, many tasks can be undertaken. For example, we can test and improve the effectiveness of goal interpretation on more benchmarks such as the RoboCup@Home competition. We can enhance the OBTEA's planning efficiency using heuristics like modern planners. Last but not least, we can integrate rule discovery algorithms to automatically learn environment transition rules from experience, thereby enhancing robots' capabilities for rapidly adapting and planning BTs in unseen scenarios.

\section*{Acknowledgments}
This work was supported by the National Natural Science
Foundation of China (Grant Nos. 62106278, 91948303-1, 
611803375, 12002380, 62101575), and the National Key
R\&D Program of China (Grant No. 2021ZD0140301). The authors also thank Dataa Robotics for organizing the competition and providing technical assistance.

\section*{Contribution Statement}
The contributions of Xinglin Chen and Yishuai Cai to this
paper were equal. Corresponding author: Minglong Li.

\bibliographystyle{named}
\bibliography{ijcai24}

\end{document}